%% file: eacl2023.tex
\pdfoutput=1

\documentclass[11pt]{article}

\usepackage{EACL2023}

\usepackage{times}
\usepackage{latexsym}

\usepackage[T1]{fontenc}

\usepackage[utf8]{inputenc}

\usepackage{microtype}

\usepackage{inconsolata}

\usepackage{graphicx}
\usepackage{caption}
\usepackage{subcaption,booktabs,multirow,graphicx}

\title{Findings of the First Workshop on\\ Simulating Conversational Intelligence in Chat}

\author{Yvette Graham \\
  ADAPT Research Centre \\
  O'Reilly Institute \\
  Trinity College Dublin \\
  \texttt{graham.yvette@gmail.com} \\ \And
  Rameez Qureshi \\
  ADAPT Research Centre \\
  O'Reilly Institute \\
  Trinity College Dublin \\
  \texttt{rameez.mrq@gmail.com} \\ \And
  Haider Khalid \\
  ADAPT Research Centre \\
  O'Reilly Institute \\
  Trinity College Dublin \\
  \texttt{haider.khalid}\\ \texttt{@adaptcentre.ie} \\ \AND
  Gerasimos Lampouras \\
  Huawei Noah's Ark Lab, \\ London, UK \\
  \texttt{gerasimos.lampouras}\\ \texttt{@huawei.com} \\\And
  Ignacio Iacobacci\\
  Huawei Noah's Ark Lab, \\ London, UK \\
  \texttt{ignacio.iacobacci}\\ \texttt{@huawei.com} \\\And
  Qun Liu \\
  Huawei Noah's Ark Lab, \\ Hong Kong, China \\
  \texttt{qun.liu@huawei.com} \\
  }

\begin{document}
\maketitle

\input{sec-00-abstract}
\input{sec-01-intro}

\input{sec-02-shared-task}

\input{sec-03-results}
\input{sec-04-conc}

\section*{Acknowledgements}
The  work presented in this paper is supported the and is supported by the Science Foundation Ireland Research Centre, ADAPT at Trinity College Dublin and part funded by our industry partner Noah's Ark Lab, Huawei under Grant Agreement No 13/RC/2106\_P2.
We additionally thank all who took part in the live human evaluation.
This work has received research ethics approval by Trinity College Dublin Research Ethics Committee (Application no. 20210603).

\input{app-00-screenshots}

\bibliography{anthology,custom}
\bibliographystyle{acl_natbib}

\end{document}

%% file: sec-00-abstract.tex
\begin{abstract}

    The aim of the workshop was to bring together experts working on open-domain dialogue research. In this speedily advancing research area many challenges still exist, such as learning information from conversations, and engaging in a realistic and convincing simulation of human intelligence and reasoning.
\textsc{sci-chat} follows previous workshops on open domain dialogue but in contrast the focus of the shared task is simulation of intelligent conversation as judged in a live human evaluation. Models aim to include the ability to follow a challenging topic over a multi-turn conversation, while positing, refuting and reasoning over arguments. 
The workshop included both a research track and shared task.
The main goal of this paper is to provide an overview of the shared task, and an in depth analysis of the shared task results following presentation at the workshop. The current paper is an extension of that made available prior to presentation of results at the workshop at EACL Malta \cite{graham-etal-2024-findings}. 
The data collected in the evaluation was made publicly available to aide future research.\footnote{ \url{https://github.com/rameez-mrq/Sci-Chat-EACL-24}} The code was also made available for the same purpose.\footnote{\url{https://github.com/rameez-mrq/sci-chat-EACL}}
\end{abstract}

%% file: sec-01-intro.tex
\section{Introduction}
Despite substantial progress in conversational AI over the past number of years and heightened attention amongst the general public, effective evaluation of such systems remains a challenge.
The ideal evaluation of dialogue models consists of measurement of performance via a large group of human users who partake in conversations with models and carry out a live evaluation post chat with each model,  reporting the successes or failures that take place.
Past attempts at live human evaluation of dialogue systems have yet to be successful, as results have either relied fully on automatic metrics known to correlate poorly with human evaluation (if at all), or discarded human evaluation as it was unfortunately deemed unreliable \cite{convai2paper}. Challenges encountered were likely due to 
excessively large number of potential good responses that can be generated (causing reference-based evaluation to vastly under-reward many systems), in addition to the challenges of evaluating the many facets of human conversation that enable simulation of intelligence. Open domain dialogue subsequently provides what we consider to be one of the most challenging evaluation tasks in NLP.
In this shared task, we revisit live human evaluation of models, and apply methods proven successful in distinct NLP tasks to open domain dialogue.

%% file: sec-02-shared-task.tex
\section{Shared Task}

\begin{table*}
\centering
\footnotesize
\input{tbl-00-results}
\caption{Results of Direct Assessment human evaluation in which 0--100 ratings were collected for each of the above criteria in the form of asking the human judge to indicate how much they agree with a given Likert statement, with all criteria adopting the following statement \emph{The conversation with the chatbot was $<$criteria$>$.}, rating sample size for each model $=$ $1,088$, scores were converted to z-score for each worker's mean and standard deviation, before calculating overall average, a line indicates a statistical win for all the systems above over all those below it.}
\label{tab:results}
\end{table*}

The shared task has the focus of simulating any kind of intelligent conversation and participants were asked to submit an automated dialogue agent API with the aim of carrying out nuanced conversations over multiple dialogue turns, and the ability to posit, refute and reason over arguments. Participating systems were then interactively evaluated in a live human evaluation following the procedure described in \cite{ji-etal-2022-achieving}.

\subsection{Participating Models}

To promote accessibility and encourage participation, participants were permitted to use any pre-trained (or not) model and were provided a baseline model in the form of DialoGPT-Medium fine-tuned on Freakonomics\footnote{\url{https://freakonomics.com/}} podcast transcripts which are publicly available and crawled easily with scripts provided in our Git repository.\footnote{\url{https://github.com/hkmirza/EACL2024-SCI-CHAT-SharedTask/tree/main/Dataset}}
Participants are additionally permitted to use pre-trained models that are not freely accessible to the public, but to ensure fairness, participants were requested to inform organisers to identify systems for appropriate analysis of results. Subsequently, participants were permitted to use any data for system training, including the provided podcast dataset, but also other available datasets such as: Personachat \cite{zhang-etal-2018-personalizing}, Switchboard \cite{holliman92}, MultiWOZ \cite{budzianowski2020multiwozlargescalemultidomain}, amongst others.

\subsection{Human Evaluation}

 The evaluation process aims to provide valuable insight into the performance of AI in generating human-like conversation. Human assessment is used as the primary/official results of the competition, and this human evaluation is carried out using the Direct Assessment method adapted for Open-domain dialogue \cite{ji-etal-2022-achieving} that employs quality controlled crowd-sourcing, described further below. 

During human evaluation, human judges were given an assigned topic from a past podcast to discuss with models. This conversation topic is essentially from the title of a conversation within the crawled Freakonomics podcast dataset, such as \emph{``What's the point of nostalgia?"}, \emph{``Can you really manifest success through positive visualization?"}, or \emph{``New technologies always scare us. Is AI any different?"} This resulted in our aim to provide a way of steering conversations that took place in the human evaluation in the direction of an intelligent topic.\footnote{Topics were intentionally taken from within the dataset used to train systems. However, a harder test for systems would be to use topics outside the training data set. We aim to do both in future evaluations and compare results.}
After the conversation takes place via a text-based interface, the crowd-sourced human evaluator then rates the performance of the model under a number of criteria using Direct Assessment. 

\subsection{Direct Assessment}

Direct Assessment (DA) evaluation was first developed to assess the quality of machine translation output and overcomes past challenges and biases by asking evaluators to assess a single system on a continuous rating scale using Likert type statement \cite{graham-etal-2013-continuous}.
DA includes accurate quality control of crowd-sourcing and enables improvements over time to be measured \cite{graham-etal-2014-machine}, and a more accurate and cost-effective gold standard for quality estimation systems \cite{graham-etal-2016-glitters,graham-etal-2017-improving}, and has been used to train MT metrics \cite{ma-etal-2017-blend}, as well as rank systems in WMT competitions due to its ability to accurately filter out low quality hits on crowd-sourcing services such as Mturk \cite{kocmi-etal-2022-findings}.
Besides machine translation, DA has also been used to evaluate and produce official results of shared tasks in  natural language generation \cite{mille-etal-2018-first,mille-etal-2019-second,mille-etal-2020-third} and TRECVid video captioning competitions \cite{awad2023overview}.


\subsection{Participating Systems}

A total of three entries were submitted by research teams to the competition. Firstly, \textsc{SarcEmp} \cite{rizwan-2024-sarcemp}, aim to humanize the chatbot based on the user’s emotional responses and context of the conversation and  incorporate humor/sarcasm for better user engagement.
Secondly, \textsc{Emo-gen BART} \cite{debnath-etal-2024-emo}, an emotion-informed multi-task BART-based model which performs dimensional and categorical emotion detection and uses that information to augment the user inputs/responses.
Finally, the \citet{hassan-graham-2024-advancing} model is based on fine-tuning the Snorkel-Mistral-PairRM-DPO language model on podcast conversation transcripts.

Four baseline systems are included, consisting of two state-of-the-art transformer-based conversational models: \textsc{anon-b}/\textsc{anon-d}, fine-tuned on different conversational datasets, both having 345 million parameters; and two sequence-to-sequence language models: \textsc{anon-a}/\textsc{anon-c}, with 400 million and 139 million parameters respectively. All systems have been pre-trained using publicly available dialogue datasets, such that both variants of architectures has one model fine-tuned on Reddit dataset and another on PersonaChat dataset. This setup provides baselines for a variety of architectures fine tuned on distinct datasets.


\subsection{Evaluation Criteria}

Human evaluation participants were asked to state the degree to which they agree with the following statement \emph{``The conversation with the chatbot was $<$criteria$>$"}.
The evaluation criteria employed are shown in Table \ref{tab:results} alongside average scores for each model that took part in the competition and a range of publicly available models.

%% file: tbl-00-results.tex
\footnotesize
\begin{tabular}{lrrrrrrrrrr}

\toprule 
\vspace{.65cm}\\
& \multirow{-4}{*}{\rotatebox[origin=c]{45}{Overall}}
& \multirow{-4}{*}{\rotatebox[origin=c]{45}{Intelligent}}
& \multirow{-4}{*}{\rotatebox[origin=c]{45}{Interesting}}
& \multirow{-4}{*}{\rotatebox[origin=c]{45}{Informative}}
& \multirow{-4}{*}{\rotatebox[origin=c]{45}{Fluent}}
& \multirow{-4}{*}{\rotatebox[origin=c]{45}{Credible}}
& \multirow{-4}{*}{\rotatebox[origin=c]{45}{Inconsistent}}
& \multirow{-4}{*}{\rotatebox[origin=c]{45}{Incoherent}}
& \multirow{-4}{*}{\rotatebox[origin=c]{45}{Repetitive}} \\
\midrule
\textsc{anon-A}         & 0.282       & 0.216    & 0.124    & $-$0.203 & 1.764    & 0.321        & 0.334     & 0.317          & $-$0.615       \\ \hline
\textsc{anon-B}          & 0.162       & 0.055    & $-$0.005 & $-$0.325 & 1.360    & 0.070        & 0.320     & 0.318          & $-$0.494      \\
\textsc{anon-C}         & 0.097       & $-$0.125 & $-$0.266 & $-$0.529 & 1.569    & 0.083        & 0.309     & 0.326          & $-$0.589       \\
\textsc{Rizwan et al.}  & 0.045       & $-$0.108 & $-$0.105 & $-$0.495 & 1.239    & 0.023        & 0.128     & 0.079          & $-$0.399       \\
\textsc{Hassan et al.}  & 0.042       & $-$0.097 & $-$0.084 & $-$0.413 & 1.516    & $-$0.067     & $-$0.140  & $-$0.208       & $-$0.173       \\
\textsc{anon-D}         & $-$0.023    & $-$0.159 & $-$0.103 & $-$0.550 & 1.278    & $-$0.082     & $-$0.085  & $-$0.079       & $-$0.400       \\ \hline
\textsc{Debnath et al.} & $-$0.168    & $-$0.414 & $-$0.335 & $-$0.661 & 1.304    & $-$0.288     & 0.054     & $-$0.221       & $-$0.784       \\
\bottomrule
\end{tabular}


%% file: sec-03-results.tex
\section{Evaluation and Results}

\begin{table}
   \centering
   \small
\input{tbl-01-passrates}
    \caption{Workers who took part in the evaluation, those who passed quality control
    and the overall pass rate based on these figures. Note: failure of quality control did not imply rejection of hits.}
    \label{tab:passrates}
\end{table}
The evaluation of all models was carried out using a presentation of models
to human chat partners on Mechanical Turk who rated the performance of each conversation
directly after talking to that model in a blind randomized test set-up.
Each HIT accepted by an Mturk worker was comprised of all models, in a random order,
in addition to a quality control model that is known to provide bad responses, in random position in the HIT.
This low quality model provides a mechanism for checking the ability of human evaluators to
discriminate between models of distinct levels of performance.

Table \ref{tab:passrates} shows numbers of workers and hits included in the evaluation before and after quality control filtering.
Quality control filtering was applied as in \citet{ji-etal-2022-achieving}, with a significance test applied to the entire distribution of ratings provided 
by a given worker to the ratings of ordinary participating models versus the
low quality model and workers who met a threshold of $p<0.05$ were included in 
the evaluation results. 

Agreement as normally reported when a categorical scale is employed, via
Cohen's Kappa coefficient does not apply to a 0--100 rating scale.
For the continuous rating scale, we can however compare ratings belonging to individual
\emph{pairs of workers} and calculate the Pearson correlation coefficient between the two sets of scores.
Doing this for all possible pairs of workers, we can then examine the distribution of Pearson $r$ correlation coefficients for pairs of workers, as a substitute for measuring agreement via the Kappa coefficient.

\begin{figure}[ht]
    \centering
    \includegraphics[width = 0.45\textwidth]{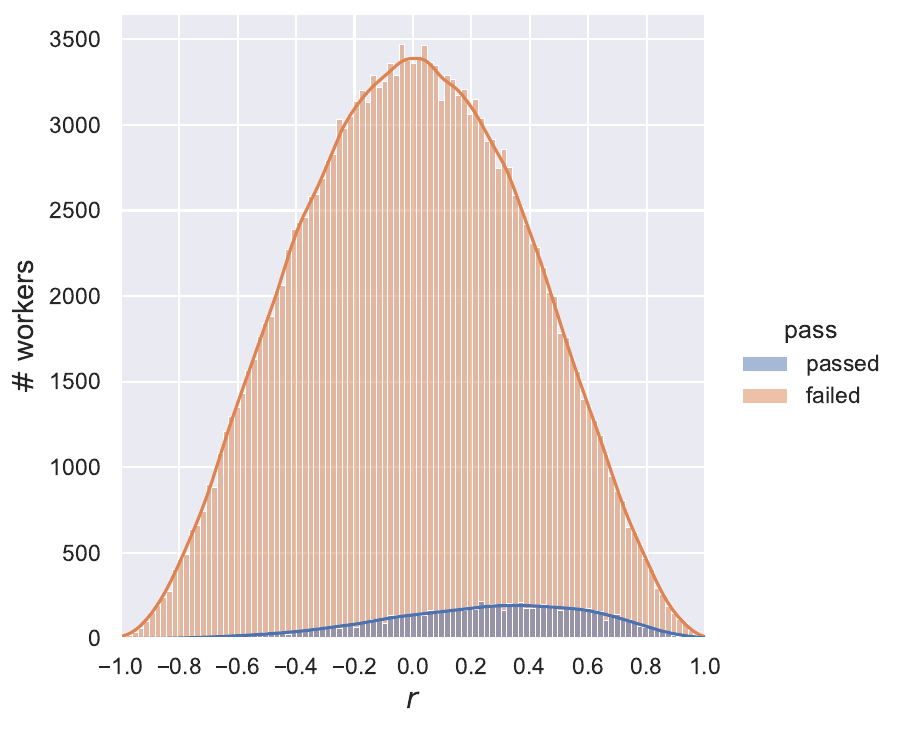}
    \caption{Rater Agreement}
    \label{fig:rater_agreement}
\end{figure}

Figure \ref{fig:rater_agreement} provides the distribution of correlation
coefficients for pairs of workers passing quality control and the same
for pairs of failing workers. 
Figure \ref{fig:topten} in Appendix \ref{appendix} additionally provides
score distributions for the ten workers who carried out the highest number of hits.

Once low quality worker hits had been removed,
score distributions for individual systems were examined, and individual criteria
ratings combined into a single overall average rating for that system. Average standardized scores are shown in Table \ref{tab:results} using method described in \citet{ji-etal-2022-achieving}.

\begin{figure}[ht]
    \centering
    \includegraphics[width = 0.45\textwidth]{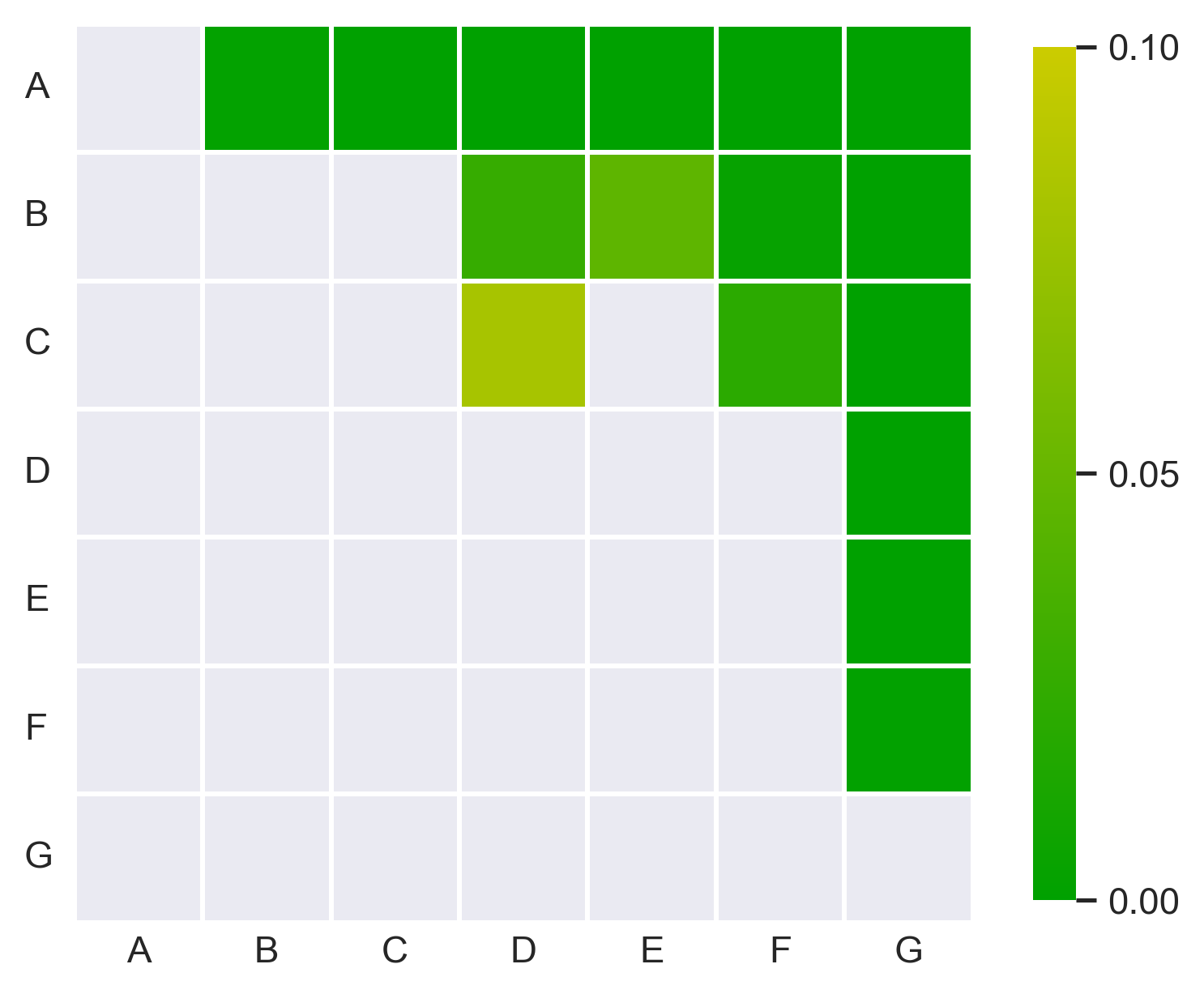}
    \caption{Significance Test results where p-value is calculated based on the distribution of ratings for each model in a Mann-Whitney U Test.}
    \label{fig:sig-test}
\end{figure}

To take into account the fact that systems with distinct overall scores can be considered a statistical tie, we apply statistical significance tests between z scores of each pair of participating system using Mann-Whitney U test, which tests the null hypothesis that the distribution of the two sets of z-scores is the same. Figure \ref{fig:sig-test} shows the results of significance test for each model pair, with a p-value below 0.05 indicating a statistical win for a system in that given row over that in the given column. A system that significantly outperforms all others is additionally indicated by a hard line in Table \ref{tab:results}.


Results of the competition showed an outright winner as \textsc{anon-a} with individual wins for pairs of systems but in terms of systems clusters, there is a large one in which we can say that all systems are technically speaking tied in a 2$^{nd}$ place cluster. Finally, a single system ranks 7$^{th}$ place as it is significantly outperformed by all others.


%% file: tbl-01-passrates.tex
\begin{tabular}{lccc}
\toprule
           & Pass QC & Total & Pass rate\\
\midrule
Workers    & 84      & 476   & 17.64\%  \\
HITs       & 134     & 763   & 17.56\%  \\
\bottomrule
\end{tabular}

%% file: sec-04-conc.tex
\section{Conclusion}

This paper describes an outline of the shared task that took place prior to the SCI-CHAT 2024 workshop to assess the ability of state-of-the-art dialogue models to simulate intelligence in conversation.
In this shared task, our aim was to apply live human evaluation to a range of publicly available chatbot models and participating systems.
To our delight, the evaluation proved successful and was feasible to carry out within a couple of weeks.
All data acquired within the context of the shared task is now public, and we hope this will provide an important resource for improving human evaluation and automatic metrics in this research area.\footnote{\url{https://github.com/rameez-mrq/Sci-Chat-EACL-24}} Code is also available.\footnote{\url{https://github.com/rameez-mrq/sci-chat-EACL}}
In future editions, we hope to advertise the competition to a larger degree to encourage a greater number of participants while keeping the evaluation manageable given resources.

%% file: app-00-screenshots.tex
\appendix
\onecolumn
\section{Appendix}
\label{appendix}

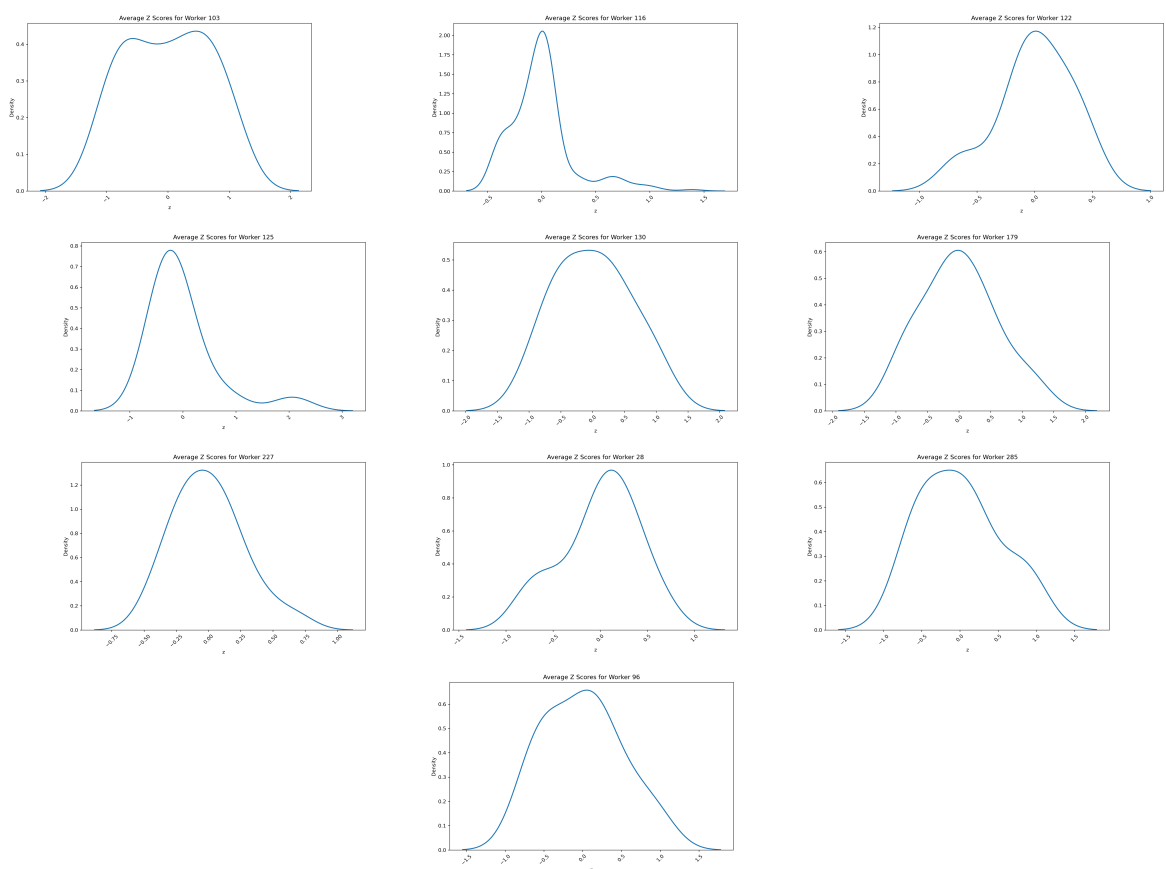
\begin{figure*}[ht]
     \centering
     \input{fig-00-rater-agree}
        \caption{Score Distributions Top 10 Individual Workers in sci-chat live human evaluation}
        \label{fig:topten}
\end{figure*}

Figure \ref{fig:topten} shows score distributions for top ten individual workers in the human evaluation who completed the highest volume of human assessments.

\begin{figure}
    \centering
    \includegraphics[width=0.75\linewidth]{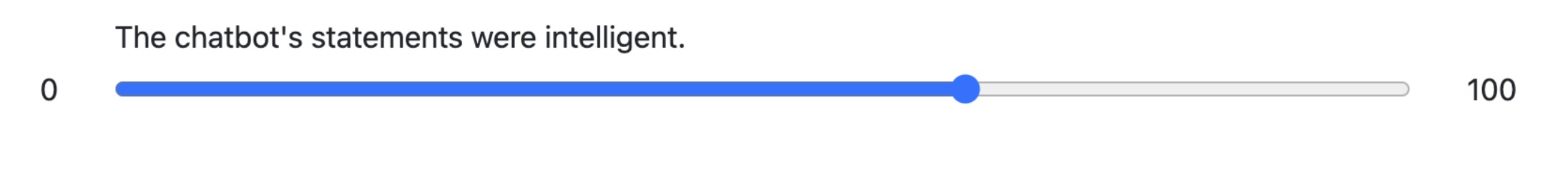}
    \caption{Continuous rating scale employed in sci-chat live human evaluation}
    \label{scale}
\end{figure}

A screenshot of the human evaluation rating scale can be found in Figure \ref{scale}.

%% file: fig-00-rater-agree.tex
\begin{subfigure}[b]{0.3\textwidth}
         \centering
         \includegraphics[width=\textwidth]{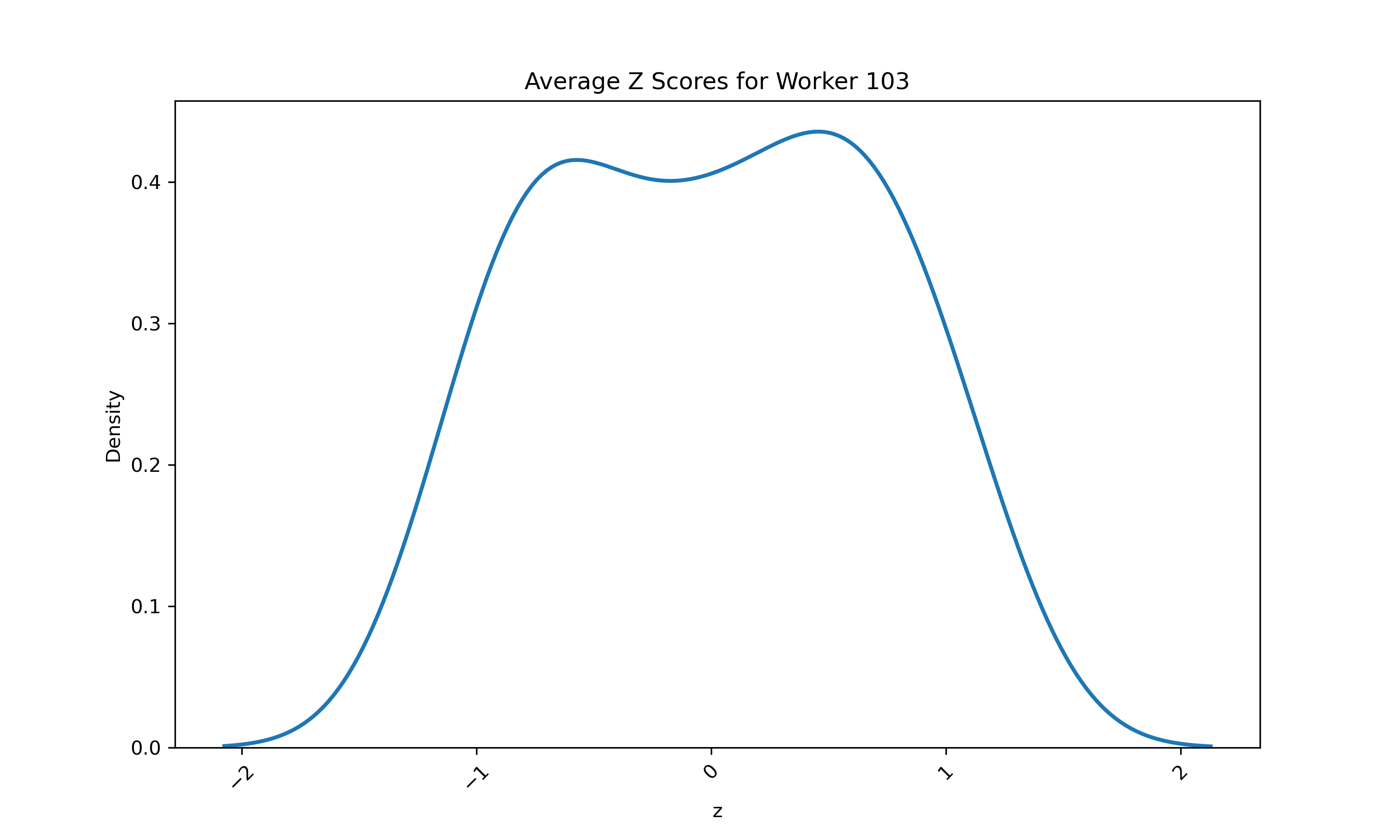}
     \end{subfigure}
     \hfill
     \begin{subfigure}[b]{0.3\textwidth}
         \centering
         \includegraphics[width=\textwidth]{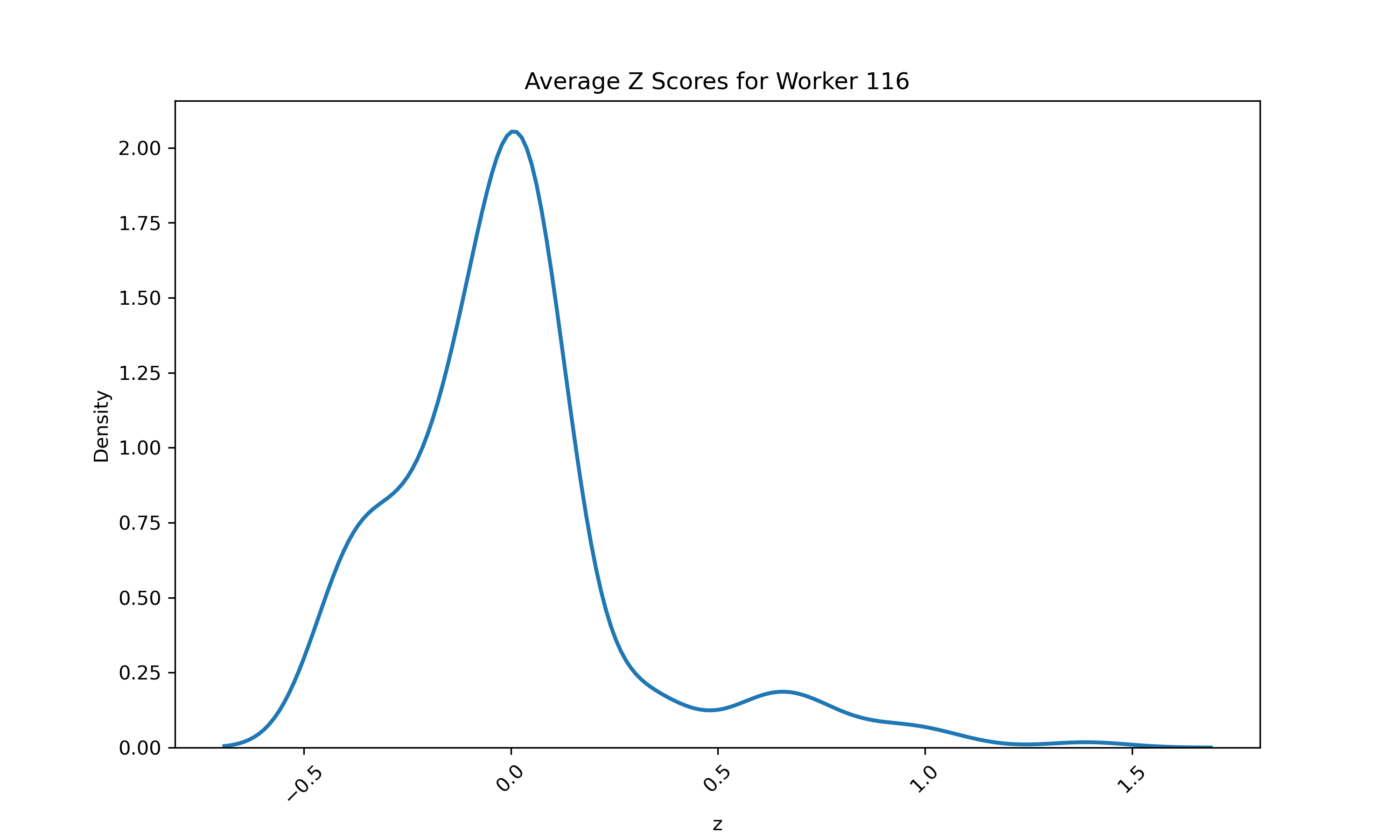}
     \end{subfigure}
     \hfill
     \begin{subfigure}[b]{0.3\textwidth}
         \centering
         \includegraphics[width=\textwidth]{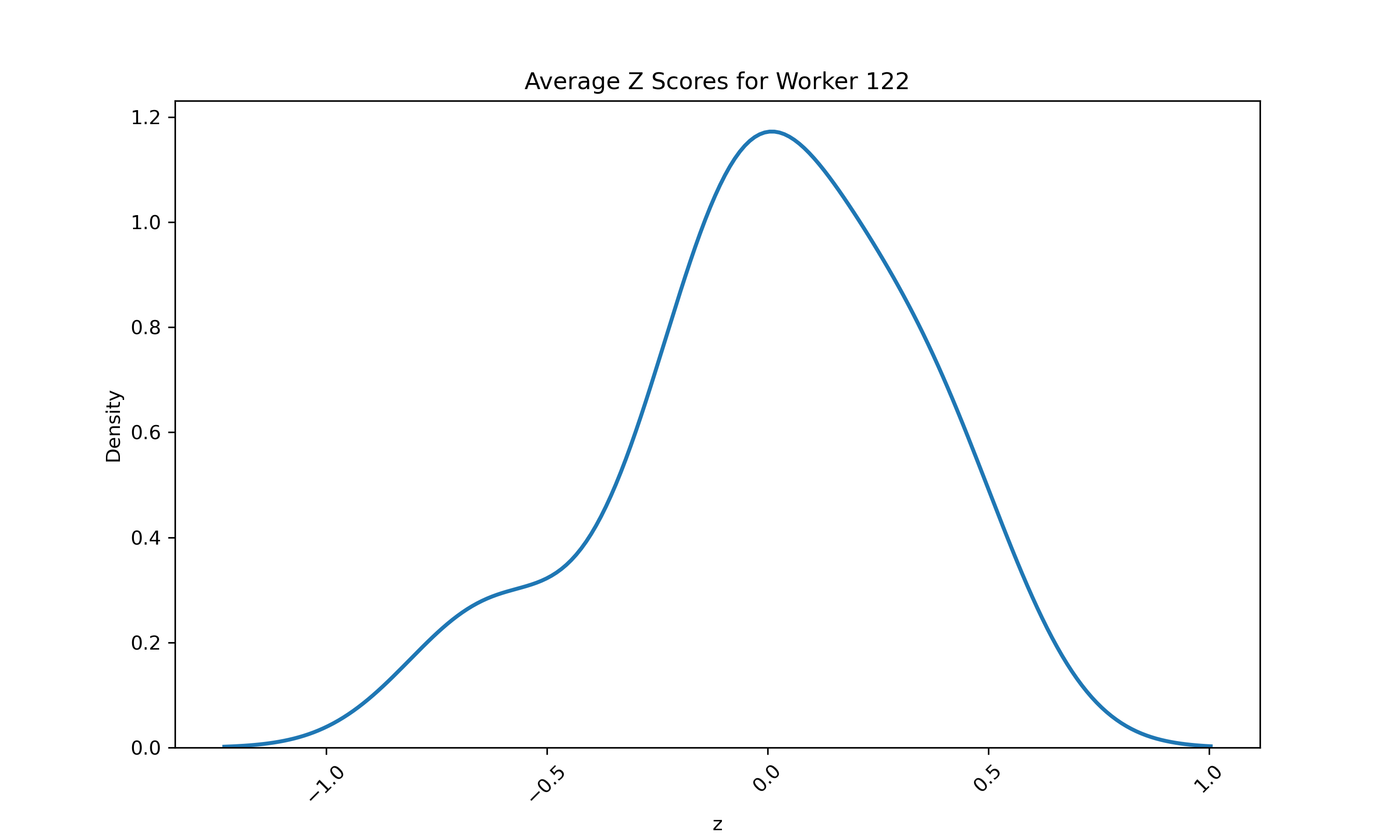}
     \end{subfigure}
      \begin{subfigure}[b]{0.3\textwidth}
         \centering
         \includegraphics[width=\textwidth]{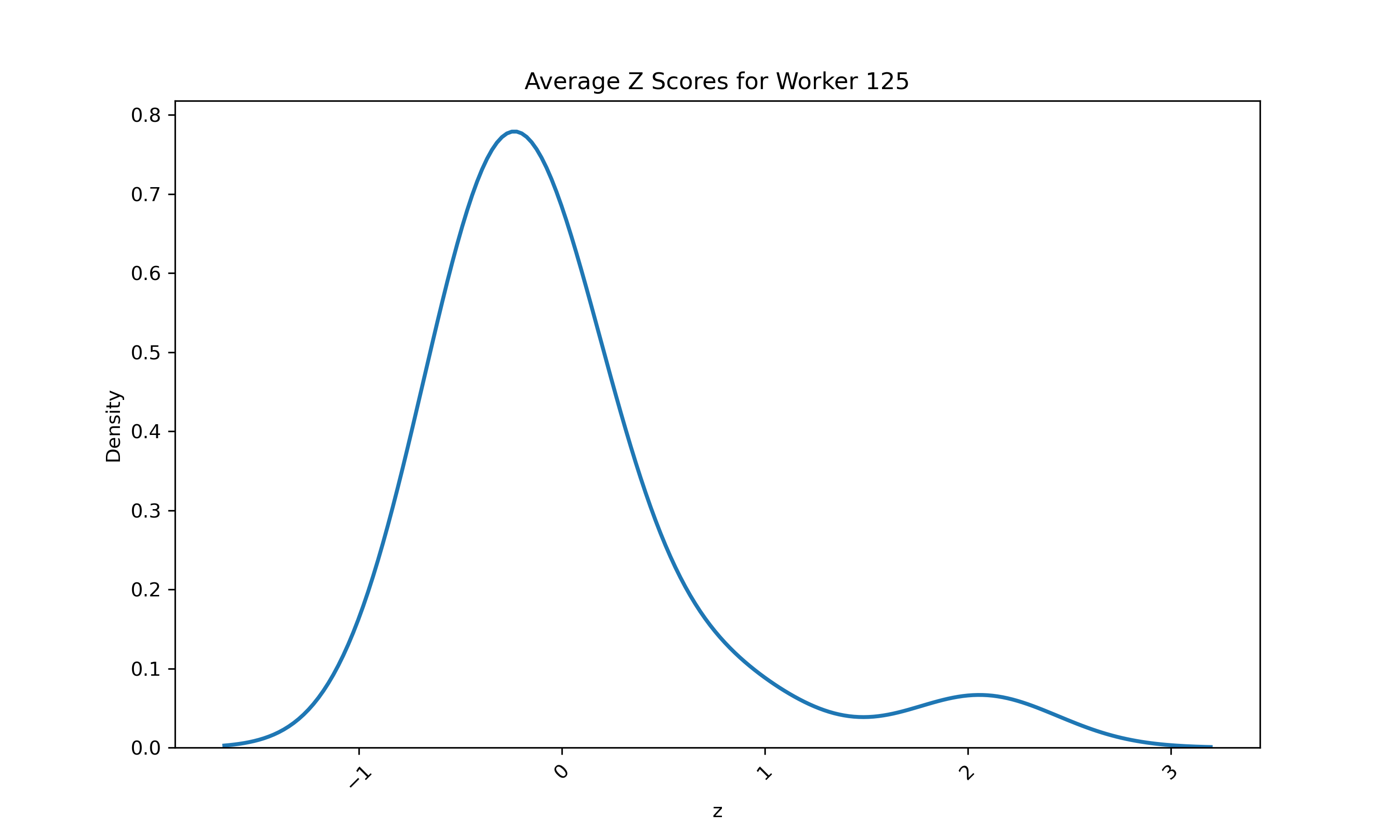}
     \end{subfigure}
      \begin{subfigure}[b]{0.3\textwidth}
         \centering
         \includegraphics[width=\textwidth]{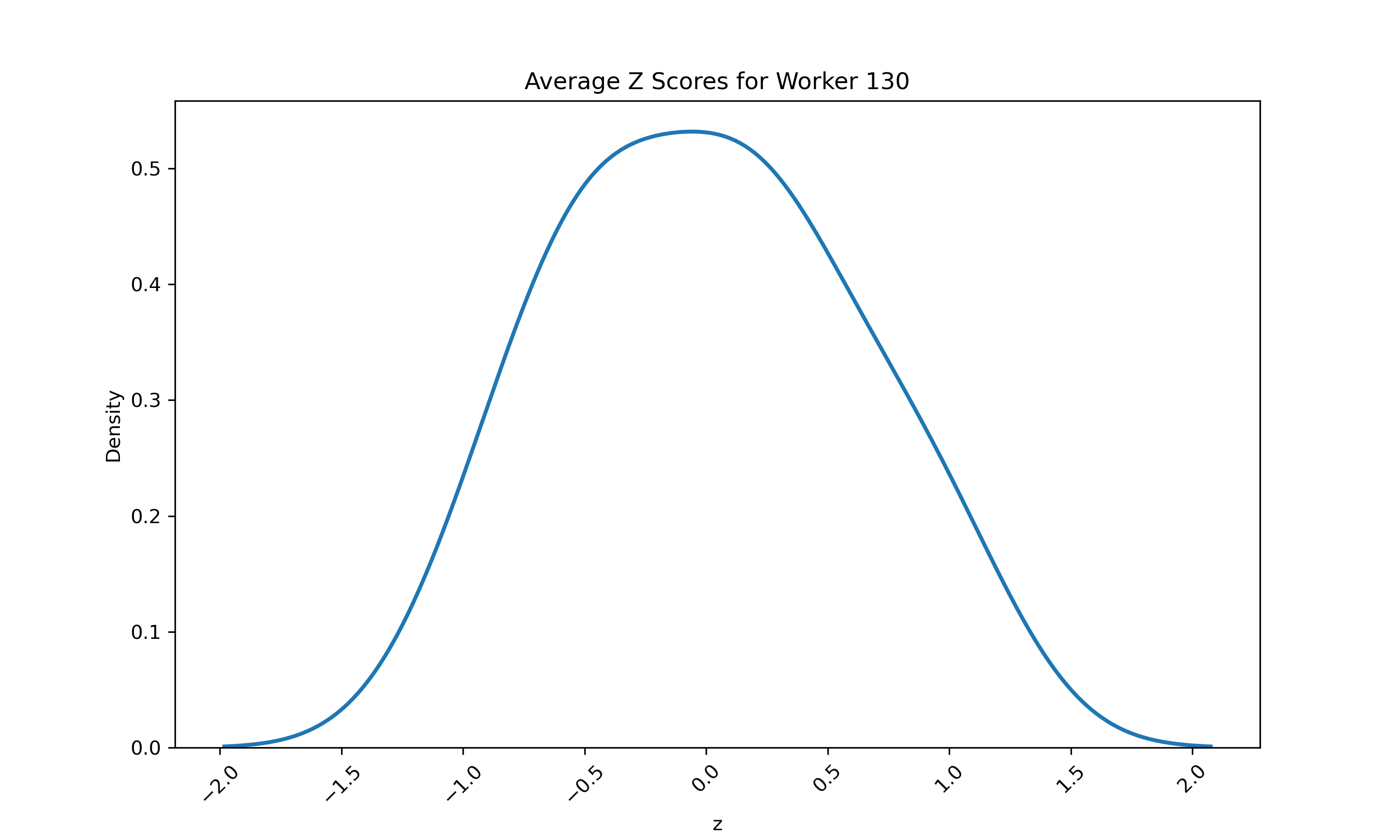}
     \end{subfigure}
      \begin{subfigure}[b]{0.3\textwidth}
         \centering
         \includegraphics[width=\textwidth]{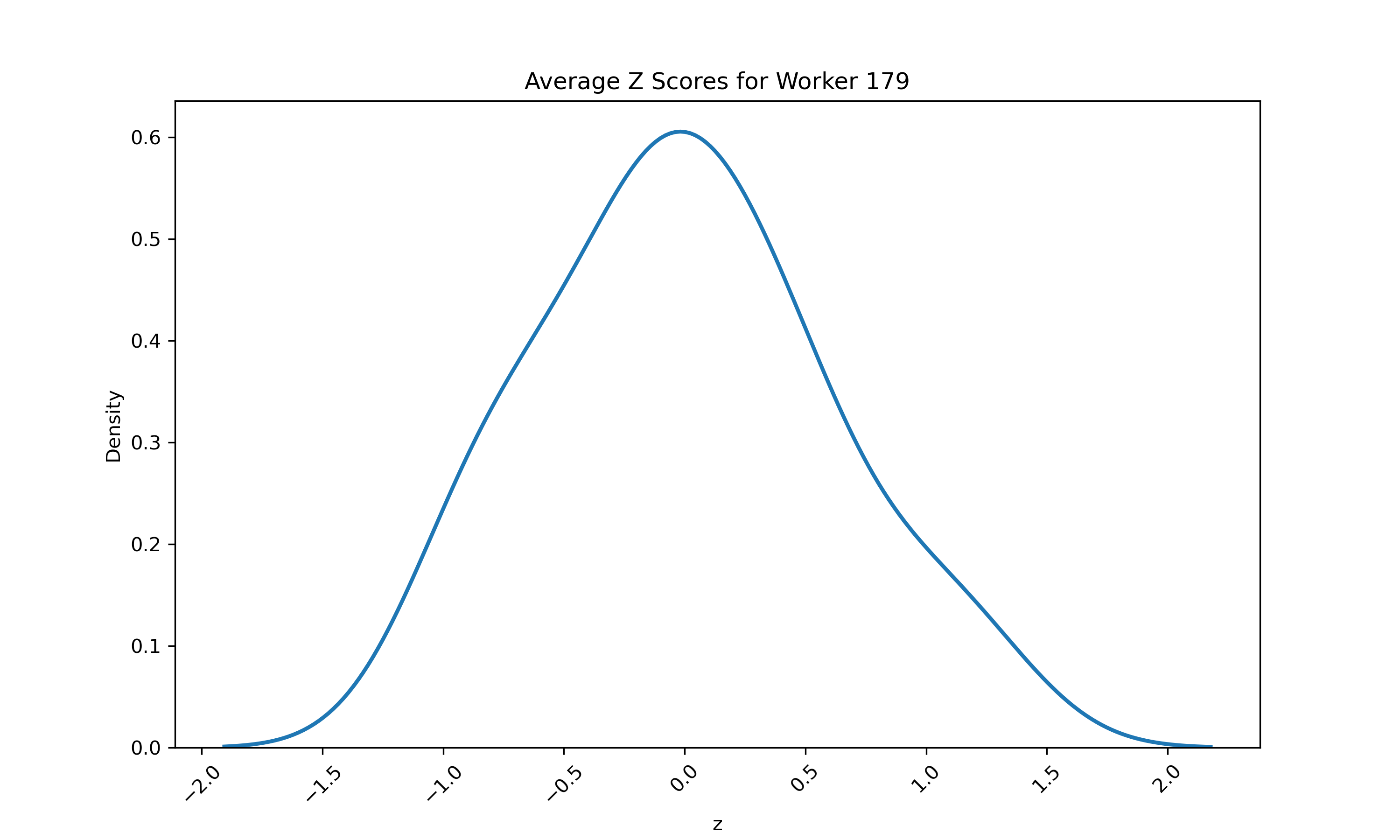}
     \end{subfigure}
      \begin{subfigure}[b]{0.3\textwidth}
         \centering
         \includegraphics[width=\textwidth]{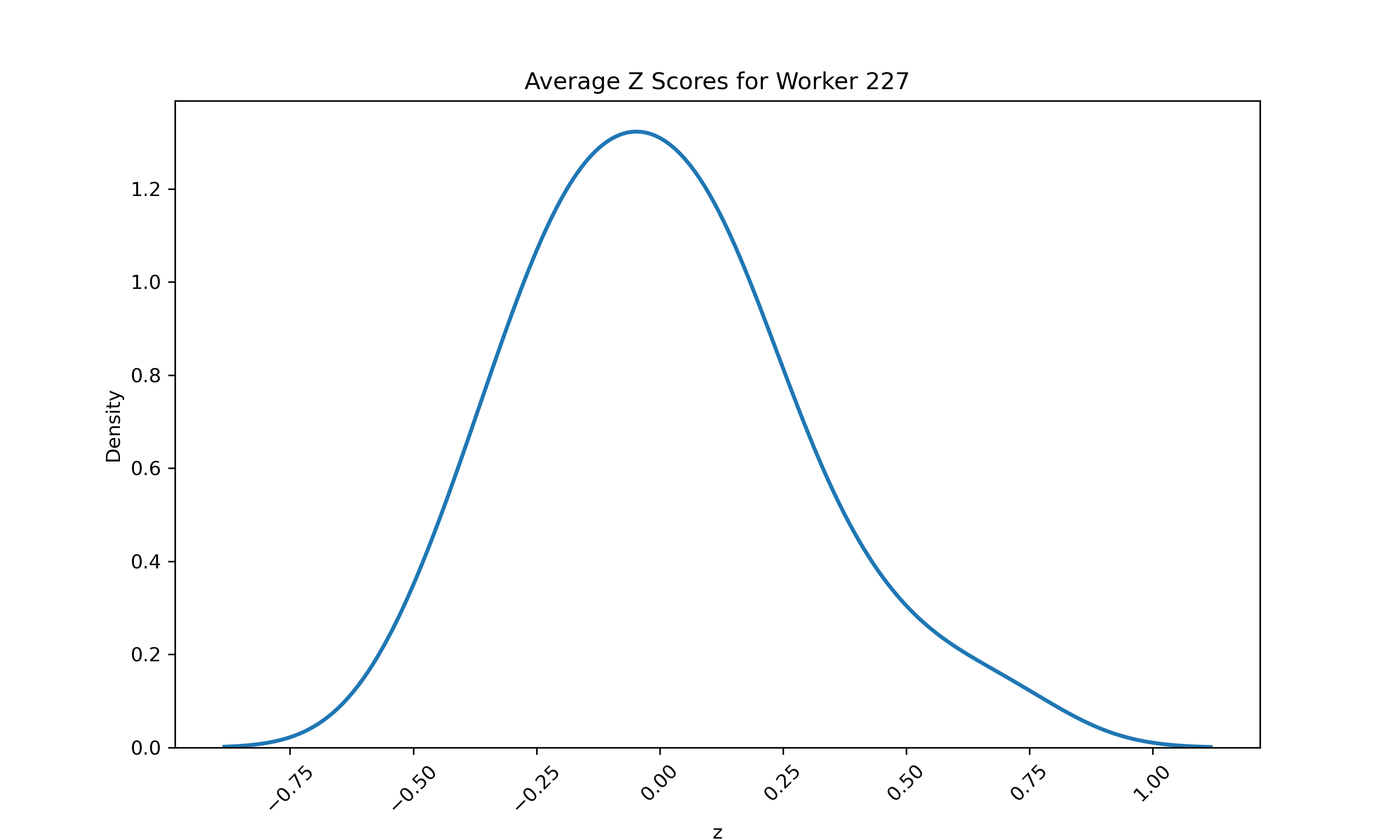}
     \end{subfigure}
      \begin{subfigure}[b]{0.3\textwidth}
         \centering
         \includegraphics[width=\textwidth]{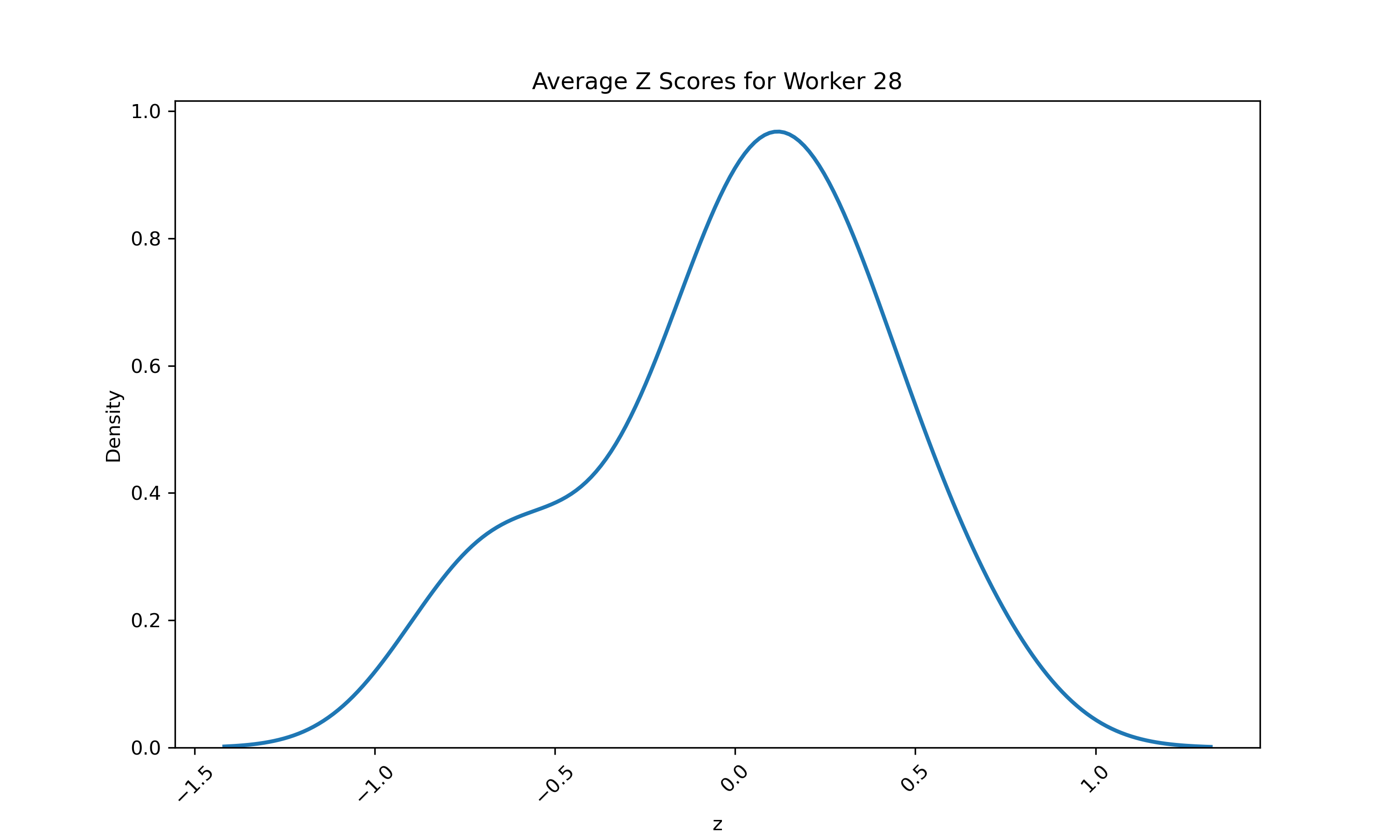}
     \end{subfigure}
      \begin{subfigure}[b]{0.3\textwidth}
         \centering
         \includegraphics[width=\textwidth]{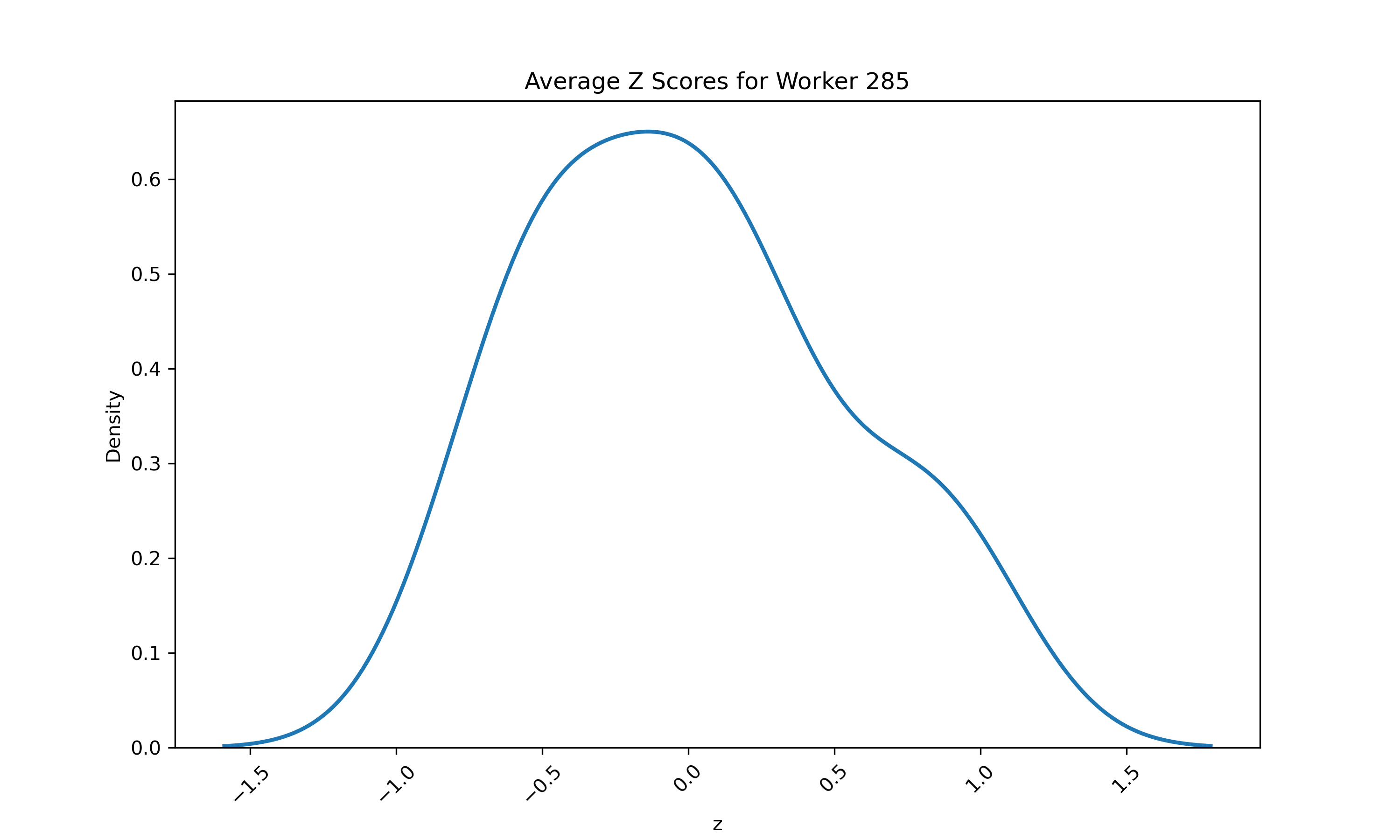}
     \end{subfigure}
      \begin{subfigure}[b]{0.3\textwidth}
         \centering
         \includegraphics[width=\textwidth]{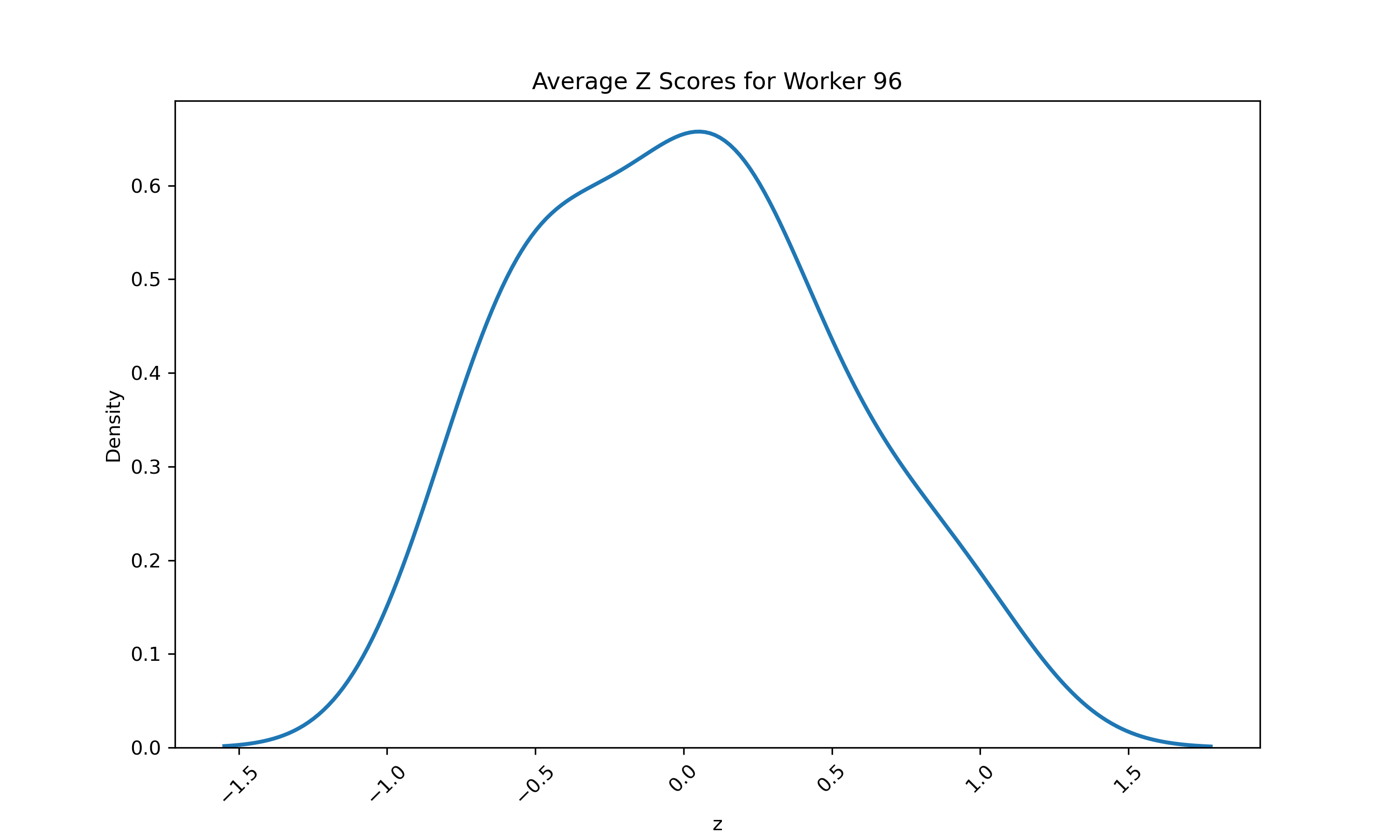}
     \end{subfigure}